\documentclass[11pt,letterpaper]{article}
\usepackage{naaclhlt2016}
\usepackage{times}
\usepackage{url}
\usepackage{alltt}
\usepackage{latexsym,graphicx,algorithmic,amsmath}
\usepackage{capt-of}
\usepackage{multirow}
\usepackage{colortbl,color}
\usepackage{booktabs}
\usepackage{algorithm}
\usepackage{enumitem}

\naaclfinalcopy 
\def\naaclpaperid{562} 

\title{Improving sentence compression by learning to predict gaze}

 \author{Sigrid Klerke \\ University of Copenhagen \\ skl@hum.ku.dk
 \And Yoav Goldberg \\  Bar-Ilan University \\ yoav.goldberg@gmail.com
 \And Anders S{\o}gaard \\ University of Copenhagen \\ soegaard@hum.ku.dk}

\date{}

\begin{document}

\maketitle

\begin{abstract}
We show how eye-tracking corpora can be used to improve sentence compression models, presenting a novel multi-task learning algorithm based on multi-layer LSTMs. We obtain performance competitive with or better than state-of-the-art approaches. 
\end{abstract}

\section{Introduction}

Sentence compression is a basic operation in text simplification which has the potential to improve statistical machine translation and automatic summarization \cite{Berg-Kirkpatrick:ea:11,Klerke:ea:15}, as well as helping poor readers in need of assistive technologies \cite{Canning:00}. This work suggests using eye-tracking recordings for improving sentence compression for text simplification systems and is motivated by two observations: (i) {\em Sentence compression is the task of automatically making sentences easier to process by shortening them.} (ii) {\em Eye-tracking measures} such as first-pass reading time and time spent on regressions, i.e., during second and later passes over the text, {\em are known to correlate with perceived text difficulty}~\cite{Rayner:12}.

These two observations recently lead Klerke et al.~\shortcite{Klerke:ea:15} to suggest using eye-tracking measures as metrics in text simplification. We go beyond this by suggesting that eye-tracking recordings can be used to induce better models for sentence compression for text simplification. Specifically, we show how to use existing eye-tracking recordings to improve the induction of Long Short-Term Memory models (LSTMs) for sentence compression.


Our proposed model \emph{does not require} that the gaze data and the compression data come from the same source.  Indeed, in this work we use gaze data from readers of the Dundee Corpus to improve sentence compression results on several datasets.  While not explored here, an intriguing potential of this work is in deriving sentence simplification models that are personalized for individual users, based on their reading behavior.

Several approaches to sentence compression have been proposed, from noisy channel models \cite{Knight:Marcu:02} over conditional random fields \cite{Elming:ea:13} to tree-to-tree machine translation models \cite{Woodsend:Lapata:11}. More recently, \newcite{Filippova:15} successfully used LSTMs for sentence compression on a large scale parallel dataset. We do not review the literature here, and only compare to \newcite{Filippova:15}. 

\paragraph{Our contributions} 
\begin{itemize}
\item We present a novel multi-task learning approach to sentence compression using labelled data for sentence compression and a disjoint eye-tracking corpus. 
\item Our method is fully competitive with state-of-the-art across three corpora. 
\item Our code is made publicly available at \url{https://bitbucket.org/soegaard/gaze-mtl16}.
\end{itemize}

\section{Gaze during reading}

Readers fixate longer at rare words, words that are semantically ambiguous, and words that are morphologically complex \cite{Rayner:12}. These are also words that are likely to be replaced with simpler ones in sentence simplification, but it is not clear that they are words that would necessarily be removed in the context of sentence compression.

\newcite{Demberg:Keller:08} show that syntactic complexity (measured as dependency locality) is also an important predictor of reading time. Phrases that are often removed in sentence compression---like fronted phrases, parentheticals, floating quantifiers, etc.---are often associated with non-local dependencies. Also, there is evidence that people are more likely to fixate on the first word in a constituent than on its second word \cite{hyona2000processing}. 
Being able to identify constituent borders is important for sentence compression, and reading fixation data may help our model learn a representation of our data that makes it easy to identify constituent boundaries.

In the experiments below, we learn models to predict the first pass duration of word fixations and the total duration of regressions to a word.
These two measures constitute a perfect separation of the total reading time of each word split between the first pass and subsequent passes.
Both measures are described below. They are both discretized into six bins as follows with only non-zero values contributing to the calculation of the standard deviation (SD):
{
\begin{tabular}{ll}
&\\
0: & measure = 0 or \\
1: & measure $<$ 1 SD below reader's average or \\
2: & measure $<$ .5 SD below reader's average or \\
3: & measure $<$ .5 above reader's average or \\
4: & measure $>$ .5 SD above reader's average or \\
5: & measure $>$ 1 SD above reader's average\\
\end{tabular}
}

	
\paragraph{First pass duration} measures the total time spent reading a word first time it is fixated, including any immediately following re-fixations of the same word. This measure correlates with word length, frequency and ambiguity because long words are likely to attract several fixations in a row unless they are particularly easily predicted or recognized. This effect arises because long words are less likely to fit inside the fovea of the eye. Note that for this measure the value 0 indicates that the word was not fixated by this reader.

\paragraph{Regression duration} measures the total time spent fixating a word after the gaze has already left it once. This measure belongs to the group of late measures, i.e., measures that are sensitive to the later cognitive processing stages including interpretation and integration of already decoded words. Since the reader by definition has already had a chance to recognize the word, regressions are associated with semantic confusion and contradiction, incongruence and syntactic complexity, as famously experienced in garden path sentences.
For this measure the value 0 indicates that the word was read at most once by this reader. 


See Table \ref{ex} for an example of first pass duration and regression duration annotations for one reader and sentence.

\begin{figure}
\begin{center}
{\small
\begin{tabular}{lcc}
\toprule
Words&{\sc First Pass}&{\sc Regressions}\\
\midrule
Are &4 &    4\\
tourists    &2& 0\\
enticed &3&0\\
by    &4&  0\\
these &2&  0\\
attractions   &3&  0\\
threatening &3&    3\\
their &5&  0\\
very &3&   3\\
existence   &3&    5\\
?      &3& 5\\
\bottomrule
\end{tabular}
}
\caption{\label{ex}Example sentence from the Dundee Corpus}
\end{center}
\end{figure}

\begin{figure}[hbt]
  \centering
  \includegraphics[width=\columnwidth]{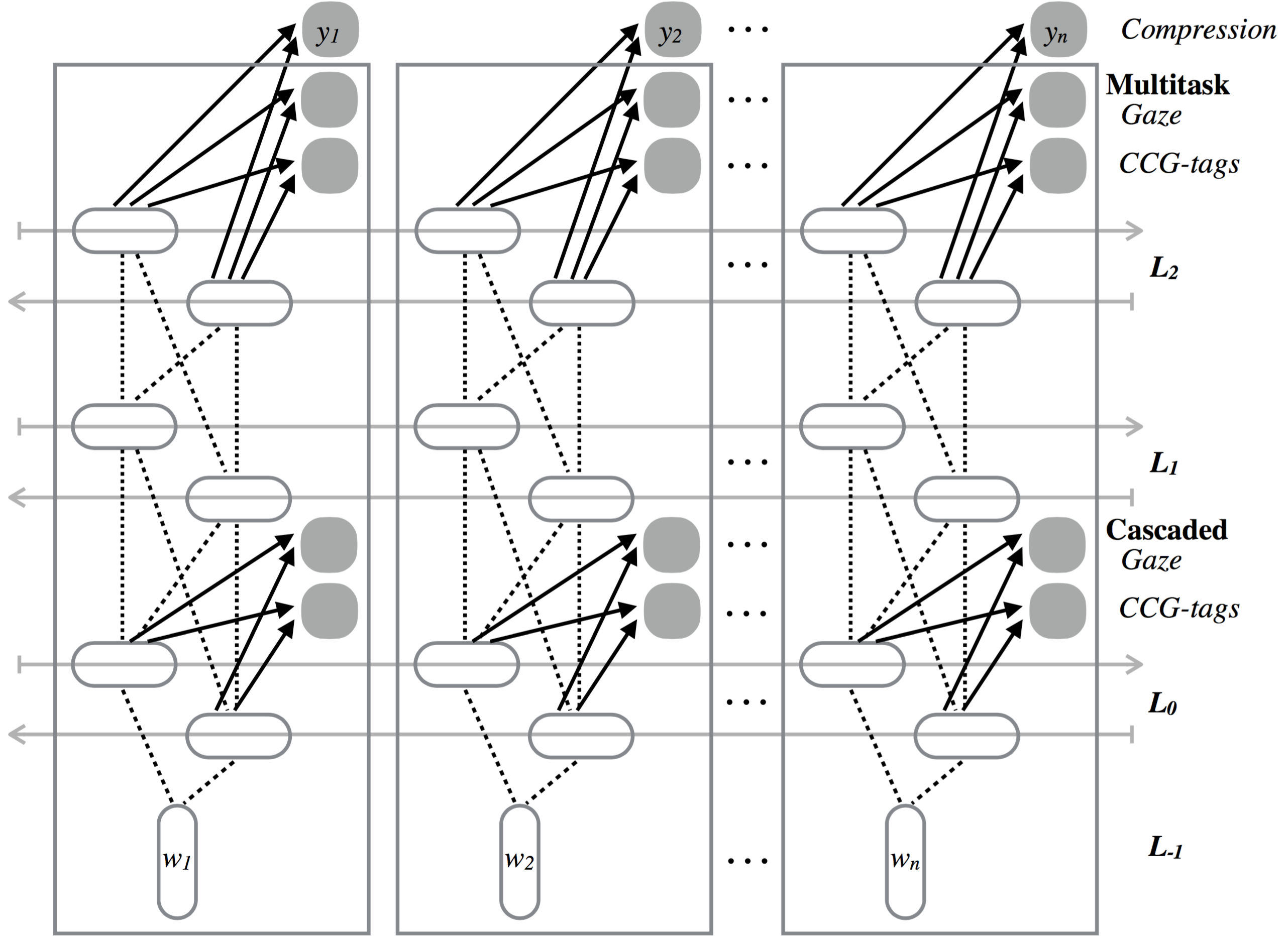}
  \caption{Multitask and cascaded bi-LSTMs for sentence compression. Layer $L_{-1}$ contain pre-trained embeddings. Gaze prediction and CCG-tag prediction are auxiliary training tasks, and loss on all tasks are propagated back to layer $L_{0}$.}
  \label{fig:model}
\end{figure}

\section{Sentence compression using multi-task deep bi-LSTMs}

Most recent approaches to sentence compression make use of syntactic analysis, either by operating directly on trees \cite{riezler2003statistical,nomoto2007discriminative,filippova2008dependency,cohn2008sentence,cohn2009sentence} or by incorporating syntactic information in their model \cite{mcdonald2006discriminative,clarke2008global}. Recently, however, \newcite{Filippova:15} presented an approach to sentence compression using LSTMs with word embeddings, but without syntactic features. We introduce a third way of using syntactic annotation by jointly learning a sequence model for predicting CCG supertags, in addition to our gaze and compression models.

Bi-directional recurrent neural networks (bi-RNNs) read in sequences in both regular and reversed order, enabling conditioning predictions on both left and right context. In the forward pass, we run the input data through an embedding layer and compute the predictions of the forward and backward states at layers $0,1,\ldots$, until we compute the softmax predictions for word $i$ based on a linear transformation of the concatenation of the of standard and reverse RNN outputs for location $i$. We then calculate the objective function derivative for the sequence using cross-entropy (logistic loss) and use backpropagation to calculate gradients and update the weights accordingly. A deep bi-RNN or $k$-layered bi-RNN is composed of $k$ bi-RNNs that feed into each other such that the output of the $i$th RNN is the input of the $i+1$th RNN. LSTMs \cite{Hochreiter:Schmidhuber:97} replace the cells of RNNs with LSTM cells, in which multiplicative gate units learn to open and close access to the error signal.

Bi-LSTMs have already been used for fine-grained sentiment analysis \cite{Liu:ea:15}, syntactic chunking \cite{Huang:ea:15}, and semantic role labeling \cite{Zhou:Xu:15}. These and other recent applications of bi-LSTMs were constructed for solving a single task in isolation, however. We instead train deep bi-LSTMs to solve additional tasks to sentence compression, namely CCG-tagging and gaze prediction, using the additional tasks to regularize our sentence compression model.
 
Specifically, we use bi-LSTMs with three layers. Our baseline model is simply this three-layered model trained to predict compressions (encoded as label sequences), and we consider two extensions thereof as illustrated in Figure~\ref{fig:model}. Our first extension, {\sc Multi-task-LSTM}, includes the gaze prediction task during training, with a separate logistic regression classifier for this purpose; and the other, {\sc Cascaded-LSTM}, predicts gaze measures from the inner layer. Our second extension, which is superior to our first, is basically a one-layer bi-LSTM for predicting reading fixations with a two-layer bi-LSTM on top for predicting sentence compressions. 
 
At each step in the training process of {\sc Multi-task-LSTM} and {\sc Cascaded-LSTM}, we choose a random task, followed by a random training instance of this task. We use the deep LSTM to predict a label sequence, suffer a loss with respect to the true labels, and update the model parameters. In {\sc Cascaded-LSTM}, the update for an instance of CCG super tagging or gaze prediction only affects the parameters of the inner LSTM layer.

Both {\sc Multi-task-LSTM} and {\sc Cascaded-LSTM} do multi-task learning \cite{Caruana:93}. In multi-task learning, the induction of a model for one task is used as a regularizer on the induction of a model for another task. Caruana~\shortcite{Caruana:93} did multi-task learning by doing parameter sharing across several deep networks, letting them share hidden layers; a technique also used by Collobert et al.~\shortcite{Collobert:ea:11} for various NLP tasks. These models train task-specific classifiers on the output of deep networks (informed by the task-specific losses). We extend their models by moving to sequence prediction and allowing the task-specific sequence models to also be deep models. 

\begin{table*}[tbb]
{\small
\begin{center}
\begin{tabular}{ll}
\toprule
{\sc S:} & \parbox[t]{14cm}{Regulators Friday shut down a small Florida bank, bringing to 119 the number of US bank failures this year amid mounting loan defaults.}\\
{\sc T:} & Regulators shut down a small Florida bank\\
\midrule
{\sc S:} & \parbox[t]{14cm}{Intel would be building car batteries, expanding its business beyond its core strength, the company said in a statement.}\\
{\sc T:} & Intel would be building car batteries\\
\bottomrule
\end{tabular}
\caption{Example compressions from the {\sc Google} dataset. S is the source sentence, and T is the target compression.}
\label{tab:examples}
\end{center}}
\end{table*}

\begin{table*}[tbh]
{\small
\begin{center}
\begin{tabular}{l|cccc}
\toprule
&Sents&Sent.len&Type/token&Del.rate\\
\midrule
\multicolumn{5}{c}{\sc Training}\\
\midrule
{\sc Ziff-Davis}&1000&20&0.22&0.59\\
{\sc Broadcast}&880&20&0.21&0.27\\
{\sc Google}&8000&24&0.17&0.87\\
\midrule
\multicolumn{5}{c}{\sc Test}\\
\midrule
{\sc Ziff-Davis}&32&21&0.55&0.47\\
{\sc Broadcast}&412&19&0.27&0.29\\
{\sc Google}&1000&25&0.42&0.87\\
\bottomrule
\end{tabular}
\caption{\label{data}Dataset characteristics. Sentence length is for source sentences.}
\end{center}}
\end{table*}

\section{Experiments}

\subsection{Gaze data}
We use the Dundee Corpus \cite{kennedy2003dundee} as our eye-tracking corpus with tokenization and measures similar to the Dundee Treebank \cite{barrett2015dundee}. The corpus contains eye-tracking recordings of ten native English-speaking subjects reading 20 newspaper articles from {\em The Independent}. 
We use data from nine subjects for training and one subject for development. We do not evaluate the gaze prediction because the task is only included as a way of regularizing the compression model.

\subsection{Compression data}

We use three different sentence compression datasets, {\sc Ziff-Davis} \cite{Knight:Marcu:02}, {\sc Broadcast} \cite{clarke2006constraint}, and the publically available subset of {\sc Google} \cite{Filippova:15}. The first two consist of manually compressed newswire text in English, while the third is built heuristically from pairs of headlines and first sentences from newswire, resulting in the most aggressive compressions, as exemplified in Table~\ref{tab:examples}. We present the dataset characteristics in Table~\ref{data}. We use the datasets as released by the authors and do not apply any additional pre-processing. The CCG supertagging data comes from CCGbank,\footnote{\url{http://groups.inf.ed.ac.uk/ccg/}} and we use sections 0-18 for training and section 19 for development.

\subsection{Baselines and system}

Both the baseline and our systems are three-layer bi-LSTM models trained for 30 iterations with pre-trained ({\sc Senna}) embeddings. The input and hidden layers are 50 dimensions, and at the output layer we predict sequences of two labels, indicating whether to delete the labeled word or not. Our baseline ({\sc Baseline-LSTM}) is a multi-task learning bi-LSTM predicting both CCG supertags and sentence compression (word deletion) at the outer layer. Our first extension is {\sc Multitask-LSTM} predicting CCG supertags, sentence compression, and reading measures from the outer layer. {\sc Cascaded-LSTM}, on the other hand, predicts CCG supertags and reading measures from the initial layer, and sentence compression at the outer layer.

\begin{table*}
\begin{center}
\small
\begin{tabular}{ll|ccccc}
\toprule
{\bf LSTM} & {Gaze}&{\sc Ziff-Davis}&\multicolumn{3}{c}{\sc Broadcast}&{\sc Google}\\
\midrule
{\bf Baseline}&&0.5668&0.7386&0.7980&0.6802&0.7980\\
\midrule
\multirow{2}{*}{\bf Multitask}&{\sc FP}&0.6416&0.7413&0.8050&0.6878&0.8028\\
&{\sc Regr.}&0.7025&0.7368&0.7979&0.6708&0.8016\\
\midrule \multirow{2}{*}{\bf Cascaded}&{\sc FP}&0.6732&{\bf 0.7519}&0.8189&{\bf 0.7012}&{\bf 0.8097}\\
&{\sc Regr.}&{\bf 0.7418}&{0.7477}&{\bf 0.8217}&0.6944&0.8048\\
\bottomrule
\end{tabular}
\caption{Results ($F_1$). For all three datasets, the inclusion of gaze measures (first pass duration (FP) and regression duration (Regr.)) leads to improvements over the baseline. All models include CCG-supertagging as an auxiliary task. Note that {\sc Broadcast} was annotated by three annotators. The three columns are, from left to right, results on annotators 1--3. }
\label{tab:results}
\end{center}
\end{table*}

\subsection{Results and discussion}

Our results are presented in Table~\ref{tab:results}. We observe that across all three datasets, including all three annotations of {\sc Broadcast}, gaze features lead to improvements over our baseline 3-layer bi-LSTM. Also, {\sc Cascaded-LSTM} is consistently better than {\sc Multitask-LSTM}. Our models are fully competitive with state-of-the-art models. For example, the best model in \newcite{Elming:ea:13} achieves 0.7207 on {\sc Ziff-Davis}, \newcite{clarke2008global} achieves 0.7509 on {\sc Broadcast},\footnote{On a "randomly selected" annotator; unfortunately, they do not say which. James Clarke (p.c) does not remember which annotator they used.} and the LSTM model in \newcite{Filippova:15} achieves 0.80 on {\sc Google} with much more training data. The high numbers on the small subset of {\sc Google} reflects that newswire headlines tend to have a fairly predictable relation to the first sentence. With the harder datasets, the impact of the gaze information becomes stronger, consistently favouring the cascaded architecture, and with improvements using both first pass duration and regression duration, the late measure associated with interpretation of content. Our results indicate that multi-task learning can help us take advantage of inherently noisy human processing data across tasks and thereby maybe reduce the need for task-specific data collection.


\section*{Acknowledgments}

Yoav Goldberg was supported by the Israeli Science Foundation Grant No.~1555/15. Anders S{\o}gaard was supported by ERC Starting Grant No.~313695. Thanks to Joachim Bingel and Maria Barrett for preparing data and for helpful discussions, and to the anonymous reviewers for their suggestions for improving the paper. 


\bibliography{../biblio}
\bibliographystyle{naaclhlt2016}

\end{document}